\documentclass[10pt,twocolumn,letterpaper]{article}

\usepackage{cvpr}              % To produce the CAMERA-READY version
\usepackage{subcaption}
\usepackage{tikz}
\usepackage{makecell}
\definecolor{cvprblue}{rgb}{0.21,0.49,0.74}
\usepackage[pagebackref,breaklinks,colorlinks,allcolors=cvprblue]{hyperref}

\def\confName{CVPR}
\def\confYear{2025}

\title{LMFormer: Lane based Motion Prediction Transformer}

\author{
    Harsh Yadav$^1$ \quad Maximilian Sch\"afer$^2$ \quad Kun Zhao$^2$ \quad Tobias Meisen$^1$\\
    $^1$University of Wuppertal, Germany \quad $^2$ Aptiv Services Deutschland GmbH \\
    {\tt\small {harsh.yadav@uni-wuppertal.de}}
}

\usepackage{acro}

\DeclareAcronym{ad}{
	short=AD,
	long=Autonomous Driving,
}

\DeclareAcronym{adas}{
	short=ADAS,
	long=Advance Driver Assistance Systems,
}
\DeclareAcronym{hd}{
	short=HD,
	long=High Definition,
}
\DeclareAcronym{gnn}{
	short=GNN,
	long=Graph Neural Network,
}
\DeclareAcronym{bev}{
	short=BEV,
	long=Bird-Eye-View,
}
\DeclareAcronym{msda}{
	short=MSDA,
	long=Multi-Scale-Deformable-Attention,
}
\DeclareAcronym{cnn}{
	short=CNN,
	long=Convolutional Neural Network,
}
\DeclareAcronym{GAN}{
	short=GAN,
	long=Generative Adversial Network,
}
\DeclareAcronym{VAE}{
	short=VAE,
	long=Variational Autoencoder,
}
\DeclareAcronym{detr}{
	short=DETR,
	long=Detection Transformer,
}
\DeclareAcronym{dino}{
	short=DINO,
	long=DETR with Improved deNoising anchOr box,
}
\DeclareAcronym{caspformer}{
	short=CASPFormer,
	long=Context Aware Scene Prediction Transformer,
}
\DeclareAcronym{caspnet}{
	short=CASPNet,
	long=Context Aware Scene Prediction Network,
}
\DeclareAcronym{mlp}{
	short=MLP,
	long=Multi-Layer Perceptron,
}
\DeclareAcronym{cv}{
	short=CV,
	long=Computer Vision,
}
\DeclareAcronym{wta}{
	short=WTA,
	long=Winner-Takes-All,
}
\DeclareAcronym{nll}{
	short=NLL,
	long=Negative Log Likelihood,
}

\begin{document}
\maketitle
\begin{abstract}
Motion prediction plays an important role in autonomous driving. This study presents LMFormer, a lane-aware transformer network for trajectory prediction tasks. In contrast to previous studies, our work provides a simple mechanism to dynamically prioritize the lanes and shows that such a mechanism introduces explainability into the learning behavior of the network. Additionally, LMFormer uses the lane connection information at intersections, lane merges, and lane splits, in order to learn long-range dependency in lane structure. Moreover, we also address the issue of refining the predicted trajectories and propose an efficient method for iterative refinement through stacked transformer layers. For benchmarking, we evaluate LMFormer on the nuScenes dataset and demonstrate that it achieves SOTA performance across multiple metrics. Furthermore, the Deep Scenario dataset is used to not only illustrate cross-dataset network performance but also the unification capabilities of LMFormer to train on multiple datasets and achieve better performance.
\end{abstract}    
\section{Introduction}\label{section:introduction}
As stated by Hu et al. a typical autonomous driving pipeline involves \textit{perception, prediction}, and \textit{planning} \cite{hu2023planning} modules. Here the prediction module aims to predict the future trajectories of the agents (e.g. vehicles, bicycles, pedestrians, etc.) in an environment. The inputs to this module are the dynamic context, i.e., past trajectories of the agents, and the surrounding static context, e.g., traffic lights, road boundaries, lanes, pedestrian crossings, traffic signs, parked vehicles, construction sites, etc. The downstream planning module is tasked with identifying the optimal trajectories and generating vehicle control commands. 

% Earlier approaches \cite{cui2019multimodal,chai2019multipath} preprocess static and dynamic contexts into a rasterized \ac{bev} grid, leveraging \acp{cnn} to predict future trajectories. Subsequent works by Gai et al. and Liang et al. \cite{liang2020learning,gao2020vectornet} introduce the concept of representing static context from HD maps as a graph structure. This road graph, defined by sequences of points along center lines in vector format, eliminates the need for rasterized representations. Building on this foundation, several studies \cite{liu2021multimodal,ngiam2021scene,zhou2022hivt,zhou2023query,cheng2023forecast,lan2023sept,seff2023motionlm} adopt Transformer-based architectures, achieving state-of-the-art prediction performance on widely used benchmark datasets.

Recent advances in trajectory prediction have seen significant progress through the integration of machine learning approaches \cite{cui2019multimodal,gao2020vectornet,ngiam2021scene,zhou2023query,lan2023sept,seff2023motionlm,sun2024semanticformer,liu2024laformer,yadav2025caspformer}. However, current state-of-the-art methods still face substantial challenges, leaving considerable room for improvement before trajectory prediction can be considered a solved problem. In this study, we address some of these challenges and propose novel solutions to tackle them. Our investigation primarily focuses on three critical aspects: 

\noindent(1) Dauner et al. \cite{dauner2023parting} demonstrate that by following the road lane, their open-loop planner for the target vehicle outperforms other state-of-the-art (SOTA) methods. Since open-loop planning and marginal trajectory prediction problems are correlated, we hypothesize that a similar phenomenon should also be observed for the trajectory prediction task. Indeed, in the trajectory prediction domain, some recent studies \cite{kim2021lapred,wang2022ltp,liu2024laformer} have shown that prediction modules designed solely based on lane information as a part of the static context can still achieve SOTA performance. However, in these studies, lane awareness is divided into two modules: the first module performs lane attention and selects the most important lanes in the scene, while the second module uses the contextual information of the selected lanes for predicting future trajectories. In contrast, inspired by Transformer architectures, we argue that lane attention should be a part of the prediction module and should dynamically prioritize lane segments critical for predicting the corresponding trajectory and vice versa.

\begin{figure}
    \centering
    \begin{subfigure}[t]{0.3\linewidth}
        \includegraphics[height=4cm,trim={300 160 280 330},clip]{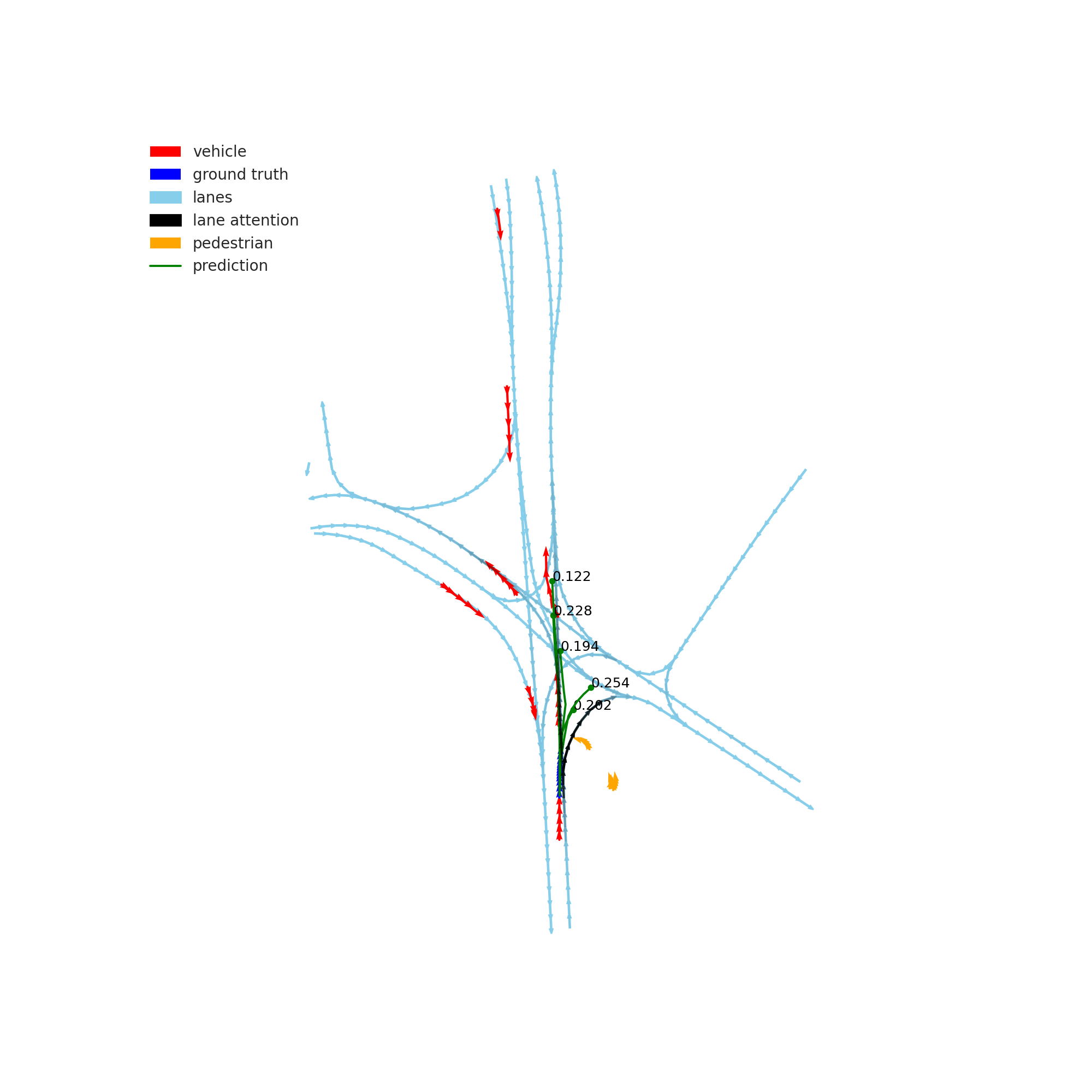}
        \caption{Initial}
    \end{subfigure}
    \begin{subfigure}[t]{0.3\linewidth}
        \includegraphics[height=4cm,trim={300 160 280 330},clip]{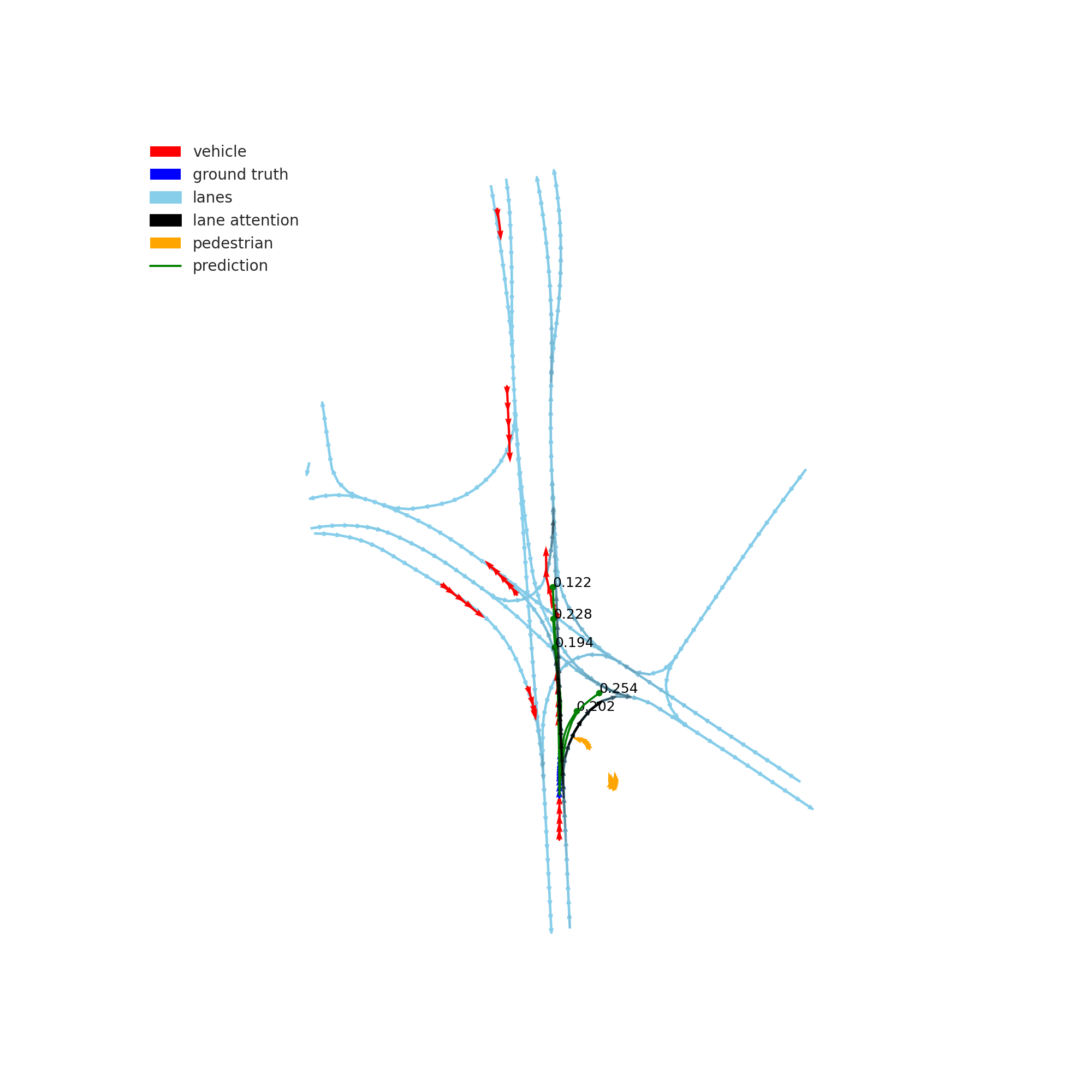}
        \caption{Refinement}
    \end{subfigure}
    \begin{subfigure}[t]{0.3\linewidth}
        \includegraphics[height=4cm,trim={300 160 280 330},clip]{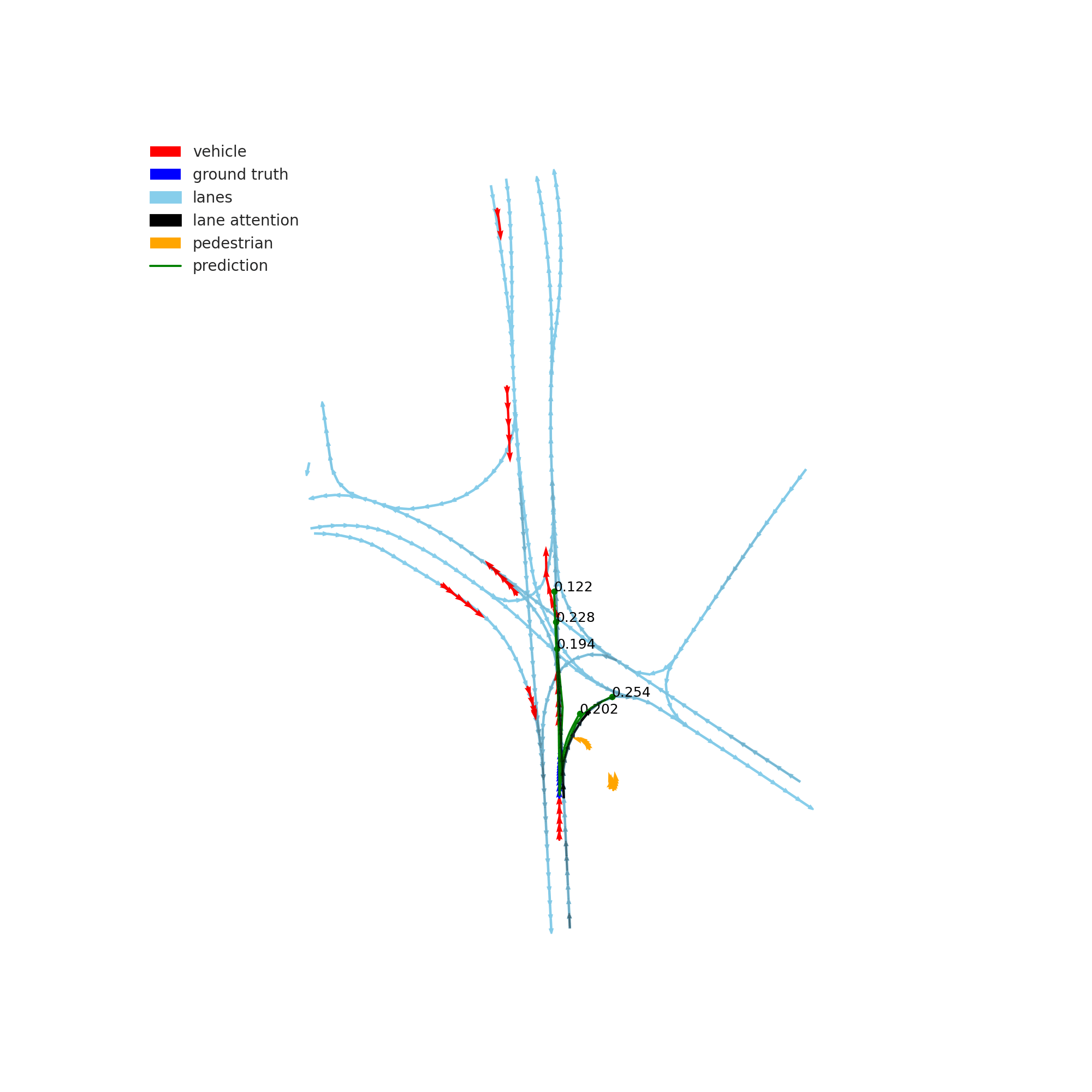}
        \caption{Final}
    \end{subfigure}
    \begin{tikzpicture}[remember picture,overlay]
     \node at (0, 3.3)      {\includegraphics[width=0.23\linewidth,trim={110 500 450 85},clip]{images_results/2706cc4eb61844a1a5c7a5cb766ffc2e_37ab2d88d07846ef96d5227e9c5a15d7_minADE5_5.01_0-min.png}};
    \end{tikzpicture}
    \caption{A schematic of predicted trajectories (green) with mode probabilities for the target agent. In (a), the initial predictions exhibit noticeable misalignment with the drivable lanes. However, through iterative refinement steps, the trajectories become increasingly structured, as seen in (b) and (c). Notably, the final predictions not only align well with the road topology but also exhibit a realistic distribution of possible future paths. }
    \label{fig:nuScenes_qual}    
\end{figure}

\noindent(2) Prior studies \cite{gao2020vectornet,kim2021lapred,zhou2022hivt,wang2022ltp,liu2024laformer,ngiam2021scene,zhou2023query} fail to account for lane connection information among different lanes at intersections, lane splits and lane merges. Sun et al. \cite{sun2024semanticformer} attempt to address this limitation by generating a knowledge graph of lane segments to learn their long-range dependency. However, this approach heavily relies on map ontology, which restricts its scalability. 

\noindent(3) Predicted trajectories frequently exhibit jerky behavior and poor lane alignment. Previous studies \cite{wang2022ltp,sun2024semanticformer} address this issue by generating a large number of trajectory proposals and filtering out those deemed unreasonable. Alternatively, other works \cite{zhou2023query,liu2024laformer} employ two-stage prediction frameworks, where the first stage produces coarse trajectory anchors, and the second stage refines these anchors by calculating the corresponding offsets. However, both approaches are computationally expensive; the former due to the generation of redundant proposals and the latter due to the additional learnable parameters required for the refinement module. Zhou et al.'s adaptive refinement method \cite{zhou2024smartrefine} mitigates these costs by performing iterative refinement with minimal computational overhead. Nonetheless, its performance is highly sensitive to hyperparameter selection, limiting its robustness in real-world applications.

To address the aforementioned challenges, we propose \textbf{L}ane based \textbf{M}otion Prediction Trans\textbf{former} (LMFormer), that incorporates lane-aware attention within the cross-attention layer of the prediction module. Our method dynamically prioritizes lane segments, using them as anchors to predict future trajectories. By keeping the lengths of individual lane segments sufficiently small, we effectively capture road curvature. However, this introduces the issue of long-range dependencies across lane segments. To overcome this, we construct a graph of lane segments that captures the long-range road structure. A graph neural network (GNN)-based map encoder is employed to enrich each lane segment's representation with information from downstream segments, including lane splits and lane merges. Notably, our map encoder avoids collecting information from irrelevant or oppositely directed lane segments, as done in \cite{liang2020learning}, ensuring greater computational efficiency. We further address the challenge of cost-effective refinement by computing additional refinement losses from intermittent output queries of stacked layers. To do so, we hypothesize that the output queries of the intermittent transformer layers should map to the same latent space as the final layer. Thus, intermittent output queries can also be transformed into trajectories and compared against the ground truth. It is important to note that the weights across different stacked layers are not shared. To the best of our knowledge, no existing trajectory prediction approach leverages intermittent outputs of stacked transformer layers to compute additional refinement losses.

We validate our proposed approach through extensive experimentation on the nuScenes \cite{caesar2020nuscenes} and the DeepScenario \cite{lu2023deepscenario} datasets, achieving SOTA performance across a range of metrics. We demonstrate that:

\begin{itemize}
    \item The incorporation of lane awareness within the cross-attention layer of the transformer-based architecture enhances explainability in the prediction module.
    \item The \ac{gnn}-based architecture effectively models long-distance dependencies across lane segments and improves the accuracy of predicted trajectories.
    \item The iterative refinement strategy refines predicted trajectories with minimal additional computational cost.
    \item LMFormer can be simultaneously trained on multiple datasets to achieve better generalization capabilities.
\end{itemize}
\section{Related Work}\label{section:related_work}
The lane-aware trajectory prediction task primarily involves three important challenges: scene representation, learning a lane graph, and forcing the diversity in the outputs. We review some of the important related work for these strategies in this section.

\subsection{Scene Representation}\label{subsection:scene_rep}
Earlier works \cite{cui2019multimodal,chai2019multipath} rely on rasterized representations of the surrounding scene, where static and dynamic context information is stored across the channel dimension. These representations allow for the use of \acp{cnn} to extract meaningful features. However, rasterized representations suffer from limited resolution and include redundant pixel information, motivating the exploration of more compact scene representations. LaneGCN \cite{liang2020learning} first introduced the idea that HD maps inherently possess a graph structure, leveraging graph convolution to encode this information. Building on this, VectorNet \cite{gao2020vectornet} proposed a unified vector representation for various scene elements, including lanes, crosswalks, traffic lights, and surrounding agents. Despite these innovations, vector-based approaches face a limitation: they encode scene elements from the perspective of the target vehicle, necessitating re-normalization and re-encoding at each time step, which introduces redundant computations. QCNet \cite{zhou2023query} addresses this issue by adopting a query-centric approach. It encodes each scene element in its own coordinate frame, removing the need for re-normalization and re-encoding. Additionally, QCNet employs temporal and spatial embeddings to capture relative information across scene elements effectively. In our work, we adopt the query-centric vector-based approach to encode scene elements, as it offers lower computational complexity compared to agent-centric and rasterized representations.

\subsection{Lane Graph} \label{subsection:lane_graph}
The lane information in the static context for motion prediction tasks is typically stored as lane polylines. Each polyline is represented as $L_p^i = [P_1^i, P_2^i, ...., P_K^i]$, where $K$ denotes the number of control points in the polyline $i$, and $P_k^i$ denotes the position of these control points in the global coordinate system. Prior studies \cite{gao2020vectornet,ngiam2021scene,liu2024laformer} first transform the polylines into the target agent coordinate system before generating their corresponding encodings. On the other hand, the approaches \cite{liang2020learning,wang2022ltp,zhou2022hivt,zhou2023query} first convert the point-based representation of the polylines to segment-based representation; $L_v^i = [V_1^i, V_2^i, ...., V_{K-1}^i]$, where $V_{k}^i = [P_k^i, P_{k+1}^i]$ stores the vector information of lane segments. In this case, the lane features are generated from the vectors of individual segments. The advantage of the segment-based approach is that road curvature is explicitly encoded in the lane encodings. A potential limitation of all the aforementioned approaches is that they fail to use explicit information on lane connections at intersections, lane splits, and lane merges. 
Sun et al. \cite{sun2024semanticformer} attempt to use lane connection information to generate consistent lanes. However, their method requires an extensive map ontology, the creation and maintenance of which is expensive and thus has limited scalability. Therefore, in our work, we propose a simple mechanism to use the lane connection information together with the segment-based representation. This integrated approach not only achieves SOTA performance, but also improves the explainability of the prediction module.  

% LaneGCN \cite{liang2020learning} pioneered the idea that lanes are among the most important static context elements for motion prediction modules. However, its convolution-based architecture performs poorly in comparison to other transformer-based approaches \cite{ngiam2021scene,zhou2022hivt}. Wang et al. \cite{wang2022ltp} argue that lane segments within the static context are particularly suitable for behavioral intent modeling. However, the authors model this intent independently of the prediction task, using a three-layer MLP to score individual lane segments. Similarly, Liu et al. \cite{liu2024laformer} score the top-k potential lane segments through interaction graphs and subsequently decode trajectories using contextual information gathered from these selected segments. In parallel, Sun et al. \cite{sun2024semanticformer} propose generating Meta-Paths based solely on static context, representing paths that the target vehicle could traverse in the near future. These learned Meta-Paths are then utilized as anchors for predicting future trajectories. A common limitation of these aforementioned approaches is that the attention mechanisms used for extracting contextual information from road lanes operate independently of the prediction task. In contrast, our work incorporates lane awareness directly into the Transformer attention mechanism within the motion prediction module. This integrated approach not only achieves SOTA performance but also enhances the explainability of the prediction module.  

\subsection{Multi-Modal Prediction}\label{section:multi-modal-prediction}
To safely plan paths in complex traffic scenarios, motion prediction modules must generate multiple modes, each representing a scene-consistent trajectory. The advent of vector-based scene representations has enabled researchers to explore techniques such as \acp{VAE} \cite{casas2020implicit,cui2021lookout} and \acp{GAN} \cite{huang2020diversitygan} for producing diverse trajectory predictions. However, these approaches often suffer from mode collapse, where all predicted modes converge to a single trajectory with minor variations. In contrast, transformer architectures leveraging learnable anchors have achieved notable success in generating diverse trajectory predictions \cite{liu2021multimodal,ngiam2021scene,zhou2022hivt,zhou2023query}. Current SOTA methods build on transformer backbones and incorporate advanced techniques, including asked auto encoders \cite{cheng2023forecast,lan2023sept}, knowledge graphs \cite{sun2024semanticformer}, and language models \cite{seff2023motionlm}. These approaches have been extensively validated on open-source datasets such as nuScenes \cite{caesar2020nuscenes}, Argoverse \cite{chang2019argoverse}, Argoverse 2 \cite{wilson2023argoverse}, and Waymo \cite{mei2022waymo}. Building on this trend, UniTraj \cite{feng2024unitraj} introduces a unified framework that consolidates different datasets, network architectures, and evaluation criteria, enabling the development of larger and more generalized models.
\section{Approach}\label{section:approach}
In this section, we describe the input-output formulation and the network architecture. We further provide a mathematical illustration of the loss function. 
\begin{figure}
    \centering
    \includegraphics[width=0.9\linewidth]{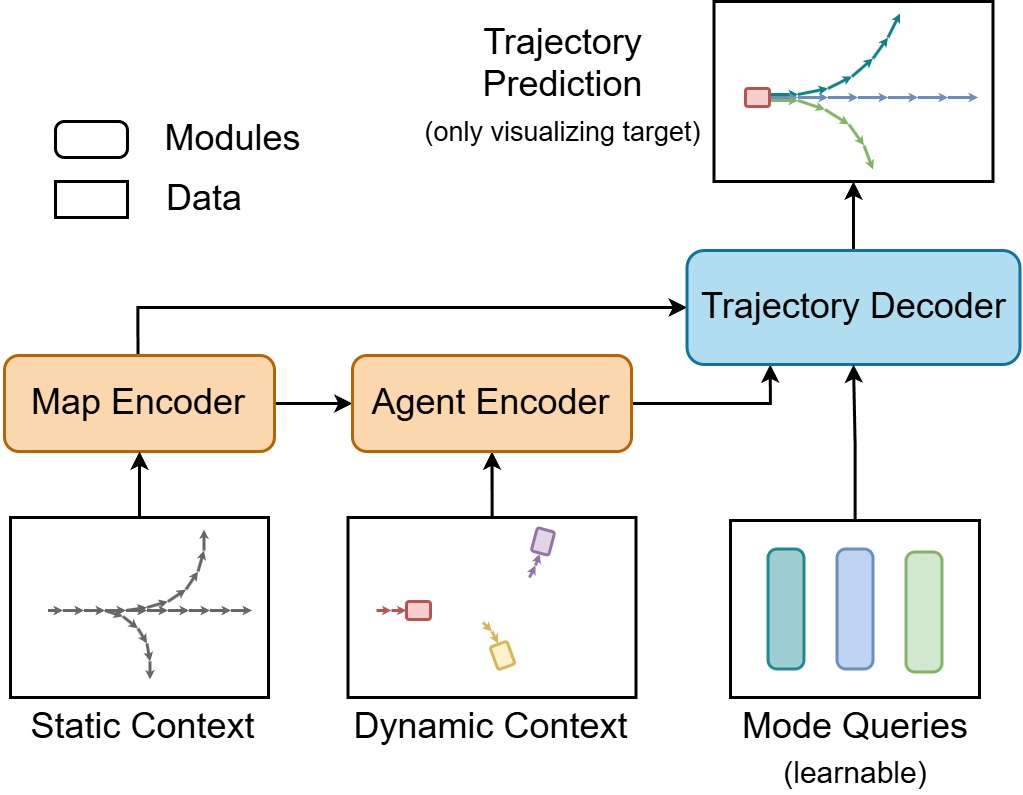}
    \caption{An Illustration of LMFormer architecture. We employ a transformer-based encoder-decoder architecture to generate multiple scene-consistent trajectories for all the dynamic agents. Notably the static context only consists of lane segments.}
    \label{fig:overall_net}
\end{figure}

\subsection{Input-Output Formulation}\label{subsection:IO_formulation}
The input to LMFormer consists of the positions of surrounding lanes and dynamic agents in the scene. As described in section \ref{subsection:scene_rep}, we opt for a query-centric representation \cite{zhou2023query} of each scene element. This representation eliminates the need for redundant agent-centric encodings at each time step, thereby reducing the computational overhead associated with repeated coordinate transformations as well as encoding generation. Below, we detail how we compute the query-centric coordinates and features for both static and dynamic contexts. In addition, we also illustrate the formulation of the output trajectory in vector format.

\noindent\textbf{Static Context:} As described in section \ref{subsection:lane_graph}, the static context consists of lane polylines from which lane segments are generated. Every lane segment is characterized by its start and end positions in global coordinates. The query-centric coordinate frame for a lane segment is set as follows: we designate the segment’s start position as its origin and align the segment's vector direction with its x-axis. As a result, the only feature preserved in this query-centric coordinate frame is the segment's length, which we use as its sole feature in the query-centric embedding generation.

\begin{figure}
    \centering
    \includegraphics[width=1\linewidth]{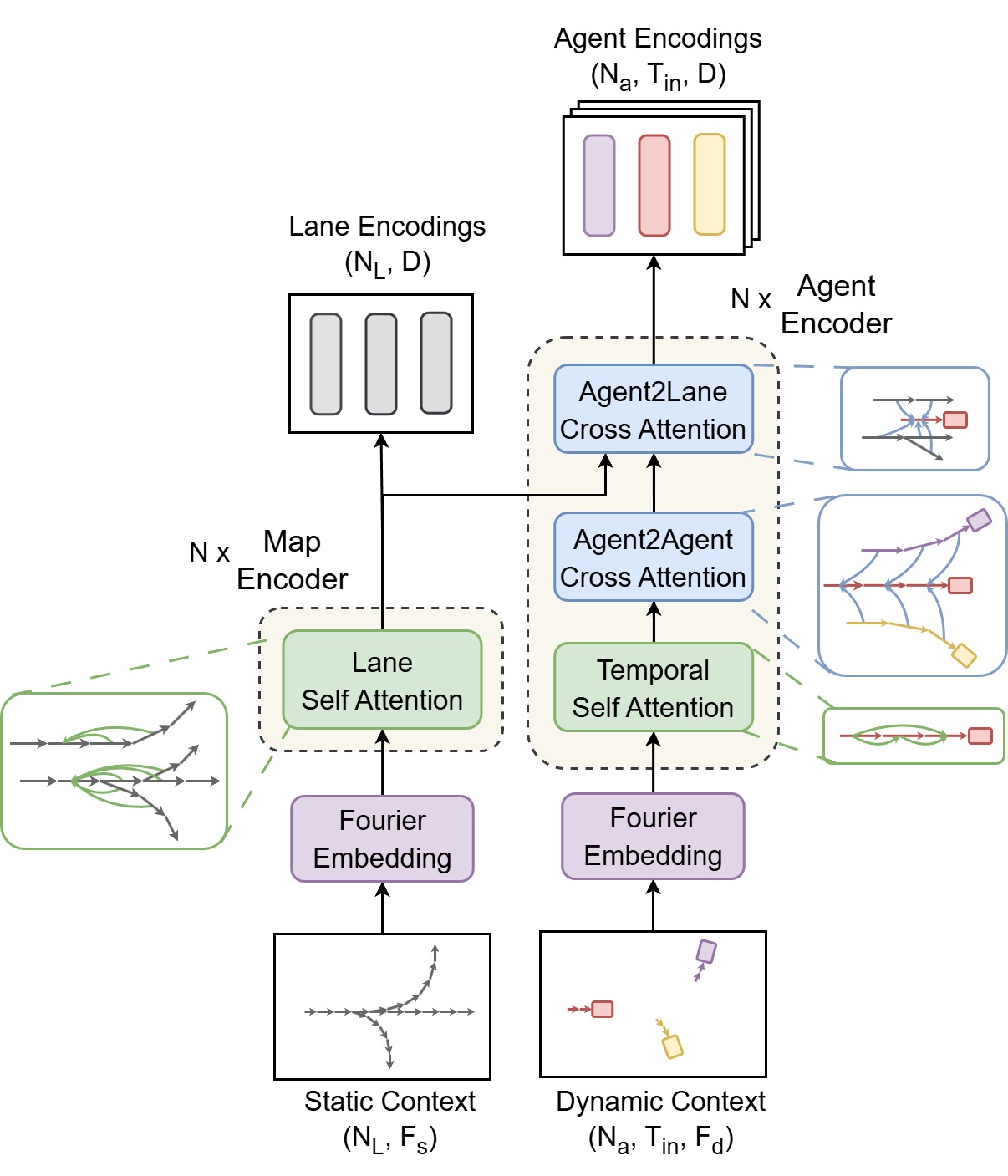}
    \caption{The encoder receives Static and Dynamic context as input and it outputs Lanes and Agents Encodings. The encoder is divided into two parts: Map Encoder and Agent Encoder. The Map Encoder models the long-range interaction among the lane segments. The Agent Encoder models the interaction of all the surrounding static and dynamic elements into each agent's latent embedding. The attention mechanisms are illustrated by green (self-attention) and blue (cross-attention) arrows, where the arrowheads point toward the queries and the tails point away from keys/values. The encoders repeat the interaction modeling N times, to learn complex interactions, where the weights across each layer are not shared.}
    \label{fig:encoder}
\end{figure}

\noindent\textbf{Dynamic Context:} For the dynamic context, agent trajectories are provided as $\mathcal{T}_{in}^a = [P_1^a, P_2^a, ..., P_T^a]$ with global positions $P_t^a$. Motion vectors $M_t^a = [P_t^a, P_{t+1}^a]$ are derived at each time step from these trajectories, encapsulating the agent’s start and end positions within that interval. The query-centric coordinate frame for a motion vector is determined as follows: we set the motion vector’s start position as its origin and align the agent’s instantaneous direction of travel with its x-axis. Notably, the motion vector’s direction may differ from the instantaneous direction of travel, introducing an angular deviation from the x-axis. Consequently, in the query-centric representation, we encode the motion vector’s length and its orientation relative to the x-axis as its features. 

\noindent\textbf{Output Representation:} The output can be represented as  $\mathcal{T}_{out}^{a,m} = [(V_1^{a,m}, S_1^{a,m}), (V_2^{a,m}, S_2^{a,m}), ..., (V_{T'}^{a,m}, S_{T'}^{a,m})]$, where $T'$ is the temporal length of the future trajectories, and $V_t^{a,m} = [P_{t-1}^{a,m}, P_t^{a,m}]$ is the motion vector corresponding to agent $a$, mode $m$, and time step $t$, with an associated variance of $S_t^{a,m}$. The query-centric coordinate frame for these motion vectors is defined as the last observed position and the current direction of travel of the corresponding agent. The predicted future trajectories are reconstructed from these motion vectors in the global coordinate frame, preserving mean positions and associated variances at each time step. Additionally, we predict the probability distribution over predicted trajectories, quantifying the likelihood of each mode.

\subsection{Network Architecture}\label{subsection:network_architecture}
Our model follows a transformer-based encoder-decoder architecture, where the decoder generates scene-consistent trajectories from the latent scene encodings produced by the encoder, as shown in Figure \ref{fig:overall_net}.

\subsubsection{Encoder}\label{subsubsection:encoder}
The encoder is designed to capture interactions between different scene elements. Our design leverages the fact that while the static context is independent of the dynamic context, the reverse does not hold. Consequently, we first encode the static context independently before incorporating it into the dynamic context encoding process. This necessitates two distinct encoders: a Map Encoder and an Agent Encoder, as illustrated in Figure \ref{fig:encoder}. Given the high-frequency nature of motion vectors, we apply learnable Fourier Embeddings \cite{tancik2020fourier} to all scene elements prior to modeling interactions within the encoder.

The Map Encoder is responsible for capturing long-range interactions between lane segments. To construct this interaction graph, we exploit the fact that the most relevant contextual information for each lane segment is present in the lane segments accessible via traversing in the direction of travel. These interactions are modeled using multi-headed attention \cite{vaswani2017attention}, where each lane segment is represented by its query-centric feature embeddings. To capture the relative positions in the multi-headed attention, the keys/values are concatenated with relative position embeddings \cite{zhou2023query}. The Lane Self-Attention module is applied iteratively $N$ times, refining the lane encodings before passing them to the Agent Encoder and subsequent network components.

The Agent Encoder learns interactions between agents and their surrounding scene elements. It comprises three specialized attention modules:
\begin{itemize}
    \item Temporal Self-Attention: Captures dependencies across different time steps for the same agent.
    \item Agent2Agent Cross-Attention: Models interactions between different agents at a given time step.
    \item Agent2Lane Cross-Attention: Encodes contextual interactions from lane segments (keys/values) to agents (queries).
\end{itemize}
Similar to the Map Encoder, multi-headed attention with relative position embeddings is employed in the Agent Encoder. Each interaction module is applied iteratively $N$ times, producing Agent Encodings that serve as input for the trajectory decoder.

\subsubsection{Decoder}\label{subsubsection:decoder}

\begin{figure}
    \centering
    \includegraphics[width=1.\linewidth]{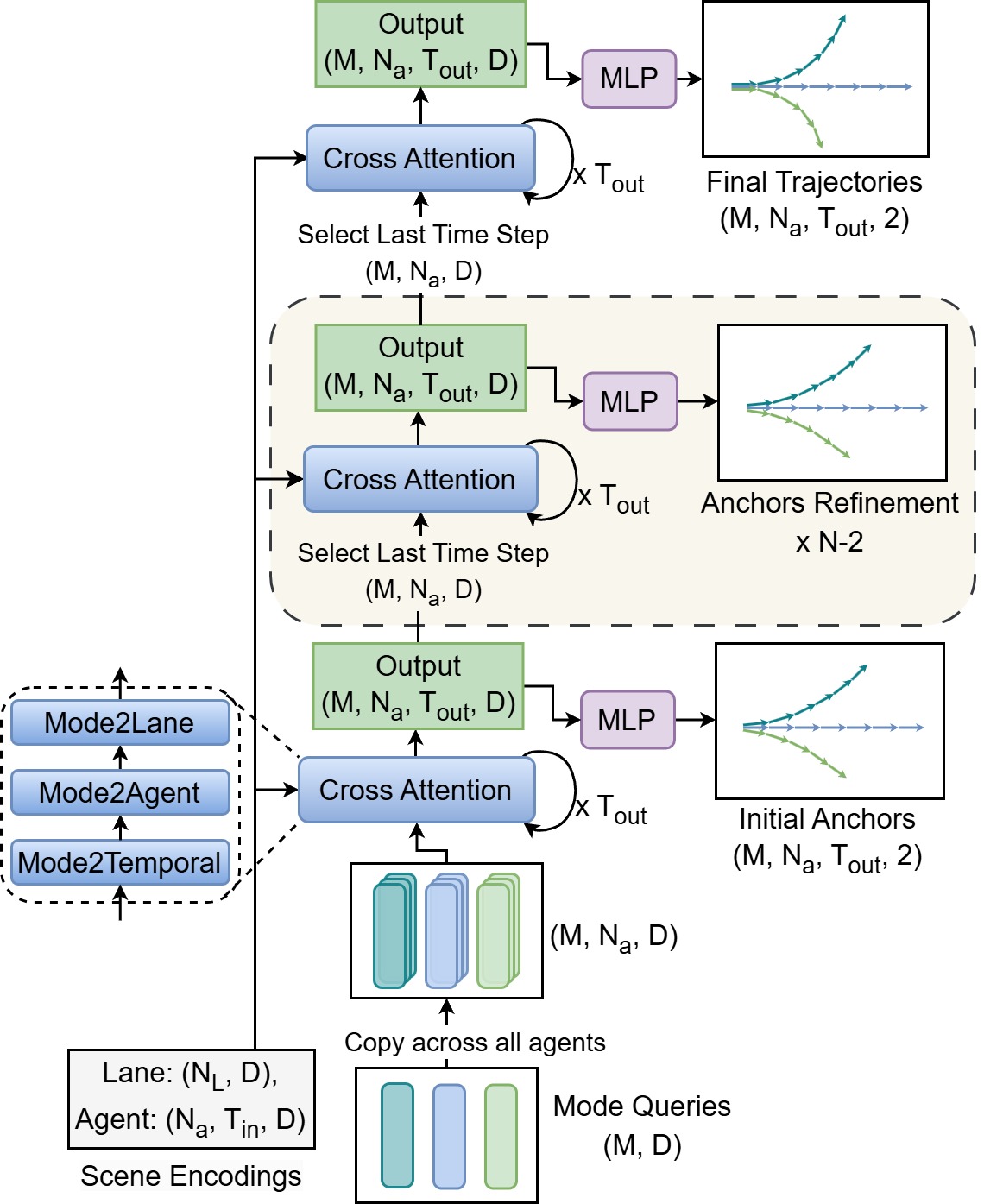}
    \caption{A depiction of the decoder architecture. The decoder performs cross-attention in between learnable mode queries and Scene Encodings (keys/values). The cross-attention layer is stacked N times and the intermittent output as well as the final output queries are transformed into trajectories with MLP. Thus we obtain N trajectories corresponding to each mode of every agent. During the training, all these N trajectories are trained against the ground truth, while during inference only the final layer output is generated. Importantly, the weights across the stacked cross-attention layers are not shared, while those in the MLP layers are.}
    \label{fig:decoder}
\end{figure}

To facilitate multimodal trajectory prediction, we introduce learnable mode queries, with each query corresponding to a distinct trajectory for every agent in the scene. Figure \ref{fig:decoder} shows the architecture of the decoder.

During initial experiments, we observed that the predicted trajectories exhibited jerky motion and, in some cases, misalignment with lane structures. To mitigate these issues, we drew inspiration from the iterative refinement of bounding boxes in DAB-DETR \cite{liu2022dabdetr}, where bounding boxes are progressively refined through offset computations after each transformer output layer. However, direct adaptation of this approach is impractical for trajectory prediction, as bounding boxes in computer vision are randomly initialized, whereas trajectory prediction requires a reliable set of initial predictions. Thus, instead of applying offset-based refinement at each transformer layer, we compute entire trajectories and compare them against the ground truth to derive additional refinement losses. Another important distinction from DAB-DETR is that LMFormer does not share the weights across each refinement layer, but rather uses the intermittent output from the multiple stacked layers. We show the effectiveness of the additional refinement loss in an ablation study.

Previous research \cite{yadav2025caspformer,zhou2023query} demonstrates that recurrent temporal decoding enhances trajectory prediction quality. Therefore, we incorporate recurrent decoding ($\times T_{out}$) in cross-attention modules. Each recurrent loop updates the query position as well as the query itself, similar to CASPFormer \cite{yadav2025caspformer}. The cross-attention module consists of three key attention mechanisms:
\begin{itemize}
    \item Mode2Temporal Cross Attention: Aggregates temporal information from Agent Encodings (keys/values) to modes (queries).
    \item Mode2Agent Cross Attention: Each mode of an agent attends to the same mode of the other agent in the scene.
    \item Mode2Lane Cross Attention: Enables mode queries to incorporate static context information from Lane Encodings (keys/values).
\end{itemize}

\subsection{Loss Formulation}\label{subsection:loss}
We model the predicted trajectories using a mixture model, which is trained to maximize the likelihood of the ground truth trajectory. The optimization objective is thus formulated as follows:

\begin{equation}
    \mathcal{L}(Y) = \sum_{m=1}^M \pi_m \prod_{t=1}^T P(Y^t | \mu_m^t, b_m^t),
    \label{eq:liklihood_function}
\end{equation}
where $Y$ represents the ground truth trajectory, $\pi_m$ denotes the probability of mode $m$, and $P(\cdot|\cdot)$ follows a Laplace distribution parameterized by the mean $\mu_m$ and scale $b_m$, which defines the positional uncertainty of the predicted trajectories. %In our initial experiments, we also tested Gaussian probability density functions as a candidate for $P(\cdot|\cdot)$, and identified that the network predicts more diverse predictions with the Laplace probability density function.

As shown by Rupprecht et al. \cite{rupprecht2017learning}, directly optimizing the \ac{nll} of mixture models can lead to numerical instability and mode collapse. To mitigate this, the mixture model optimization is decomposed into two objectives via separate regression and classification losses \cite{cui2019multimodal,makansi2019overcoming,zhou2022hivt,yadav2025caspformer}. In our approach, we adopt the \ac{wta} strategy, where the regression loss optimizes only the parameters of the best-matching mode and is defined as:

\begin{align}
    \mathrm{L}_{reg} & = - log \prod_{t=1}^T P(Y^t | \mu_{m*}^t, b_{m*}^t) \nonumber \\
                    & = - \sum_{t=1}^T log \ P(Y^t | \mu_{m*}^t, b_{m*}^t),
\end{align}
where $m^*$ is the mode with the smallest $L_2$ distance to the ground truth trajectory. For the classification loss, we optimize the \ac{nll} of the likelihood function shown in equation \eqref{eq:liklihood_function} as follows:

\begin{equation}
    \mathrm{L}_{cls} = - log \ \mathcal{L}(Y)
\end{equation}

To enhance training stability, we decouple the optimization: the classification loss updates only the mode probabilities $\pi_m$, while the regression loss optimizes the trajectory parameters $\mu_m$ and $b_m$. Finally, the total training loss is defined as the sum of classification and regression losses:

\begin{equation}
    \mathrm{L} = \lambda \ \mathrm{L}_{cls} + \sum_{n=2}^{N}\mathrm{L}_{reg},
\end{equation}
where $N$ denotes the number of stacked layers, and $\lambda$ controls the trade-off between classification and regression losses.
\section{Experiments}\label{section:experiments}
In this section, we present the experimental evaluation of LMFormer. We first describe the datasets and experimental setup, followed by a detailed quantitative and qualitative performance analysis. Our goal is to demonstrate the effectiveness of LMFormer on both standard benchmarks and real-world-inspired scenarios while also highlighting its generalization ability across different datasets.

\subsection{Experimental Setup}\label{subsection:exp_setup}
\textbf{Dataset:} As we aim to evaluate LMFormer’s performance in structured urban environments as well as more diverse intersection-heavy scenarios, we selected the nuScenes \cite{caesar2020nuscenes} and the Deep Scenario \cite{lu2023deepscenario} datasets for benchmarking. The nuScenes dataset comprises driving scenarios from Boston and Singapore, generated using sensor data of a moving vehicle in complex scenarios. The Deep Scenario dataset is built from recordings of stationary drones, placed vertically above complex traffic scenarios, e.g. intersections.

\noindent\textbf{Metrics:} We report the quantitative performance of LMFormer on the nuScenes and Deep Scenario datasets in terms of minADE\textsubscript{k}, minFDE\textsubscript{k}, MR\textsubscript{k}, and OffRoadRate. minADE\textsubscript{k} measures the average Euclidean distance between the ground truth and the best-matching trajectory among the top-$k$ predicted trajectories. A lower value indicates better alignment with the actual observed path. This metric is particularly relevant for multi-modal prediction settings, where multiple future outcomes are possible. minFDE\textsubscript{k} computes Euclidean distances between the endpoint of the ground truth and the best prediction among the top-$k$ trajectories. It indicates how well the network predicts the final goal position. MR\textsubscript{k} measures the fraction of misses, where a miss occurs when the maximum pointwise $L_2$ distance between the ground truth and predicted trajectories is larger than two meters. A small value indicates that the predicted velocity profile matches the ground truth well. Finally, the OffRoadRate metric reports the fraction of predicted trajectories outside the drivable area.

\noindent\textbf{Implementation Details:} We train LMFormer on one Nvidia A100 GPU using AdamW Optimizer \cite{loshchilov2018decoupled}. As required by nuScenes Prediction Challenge \cite{nuscene2020prediction}, the duration of the input trajectory is limited to 2 seconds and that of the output trajectory to 6 seconds, with a sampling rate of 2Hz.  Also, the maximum number of modes $k$ is set to 5, to compute minADE\textsubscript{5} and MR\textsubscript{5} for the comparison against other SOTA methods. The same configurations are extended to the Deep Scenario Dataset, as it does not specify any restrictions on the aforementioned hyperparameters. 

Both static and dynamic contexts enclose an area of 150m x 100m, where the former is along the driving direction. In order to help the model generalize for both left-hand and right-hand driving conditions, the inputs are flipped along the driving direction with a probability of 50\%. We also identify that increasing the number of stacked layers $N$ beyond 3 does not improve the network performance, and hence we set $N$ = 3 to keep the computational costs to a minimum. A similar observation is made for the length of a lane segment, and its optimum value is recorded at 3 meters. 

\subsection{Results and Discussion}\label{subsection:results}
\begin{table}[b]
    \centering
    \caption{Comparison with state-of-the-art on the nuScenes prediction challenge test split.}
    \label{tab:experiment_table}
    \resizebox{\linewidth}{!}{
    \begin{tabular}{l|c c|c|c}
    \hline
    Method & minADE\textsubscript{5}$\downarrow$ & MR\textsubscript{5}$\downarrow$ & minFDE\textsubscript{1}$\downarrow$ & OffRoad$\downarrow$ \\
    \hline
    GOHOME \cite{gilles2022gohome}  & 1.42 & 0.57 & 6.99 & 0.04 \\
    Autobot \cite{girgis2021latent}  & 1.37 & 0.62 & 8.19 & 0.02 \\
    THOMAS \cite{gilles2022thomas}  & 1.33 & 0.55 & 6.71 & 0.03 \\
    PGP \cite{deo2021multimodal}  & 1.27 & 0.52 & 7.17 & 0.03 \\
    MacFormer \cite{feng2023macformer}  & 1.21 & 0.57 & 7.50 & 0.02 \\
    LAFormer \cite{liu2024laformer}  & 1.19 & 0.48 & 6.95 & 0.02 \\
    Socialea \cite{chen2023q} & 1.18 & 0.48 & 6.77 & 0.03 \\
    FRM \cite{park2023leveraging}  & 1.18 & 0.48 & 6.59 & 0.02 \\
    CASPNet\_v2 \cite{schafer2023caspnet++} & 1.16 & 0.50  & \textbf{6.18} & \textbf{0.01} \\
    CASPFormer \cite{yadav2025caspformer} & 1.15 & 0.48 & 6.70 & \textbf{0.01} \\
    SemanticFormerR \cite{sun2024semanticformer} & \textbf{1.14} & 0.50 & 6.27 & 0.03 \\
    \hline
    LMFormer (ours) & \textbf{1.14} & \textbf{0.47} & 6.85 & \textbf{0.01} \\
    \hline
    \end{tabular}
    }
\end{table}

We provide a comparison of LMFormer against the SOTA networks on the nuScenes Motion Prediction Challenge test split \cite{nuscene2020prediction} in Table \ref{tab:experiment_table}. LMFormer achieves SOTA results across minADE\textsubscript{5}, MR\textsubscript{5} and OffRoadRate. Qualitative results are depicted in Figure \ref{fig:nuScenes_qual}. As is evident, the trajectories undergo a substantial enhancement during the refinement process. It is noteworthy that the network is capable of making diverse predictions and aligning the predicted trajectories with a high degree of accuracy with the lanes. Furthermore, the attention map provides insight into the underlying learning behaviors of LMFormer. We observe that the network pays higher attention to the lanes, which are important for predicting its future trajectory.

It is important to note that despite achieving SOTA performance across all other metrics, LMFormer does not achieve SOTA on minFDE\textsubscript{1}. We suspect that this behavior emerges due to the fact that the predicted trajectories have to align with the lanes, which is rarely the case in the real world. Such lane alignment would predominantly impact the predicted position towards the end of the trajectory compared to the initial ones. Thus, the results of this effect are more pronounced in minFDE\textsubscript{k} compared to minADE\textsubscript{k} and MR\textsubscript{k}.

\begin{figure}
    \centering
    \begin{subfigure}[t]{0.4\linewidth}
        \includegraphics[height=4cm,trim={300 140 260 360},clip]{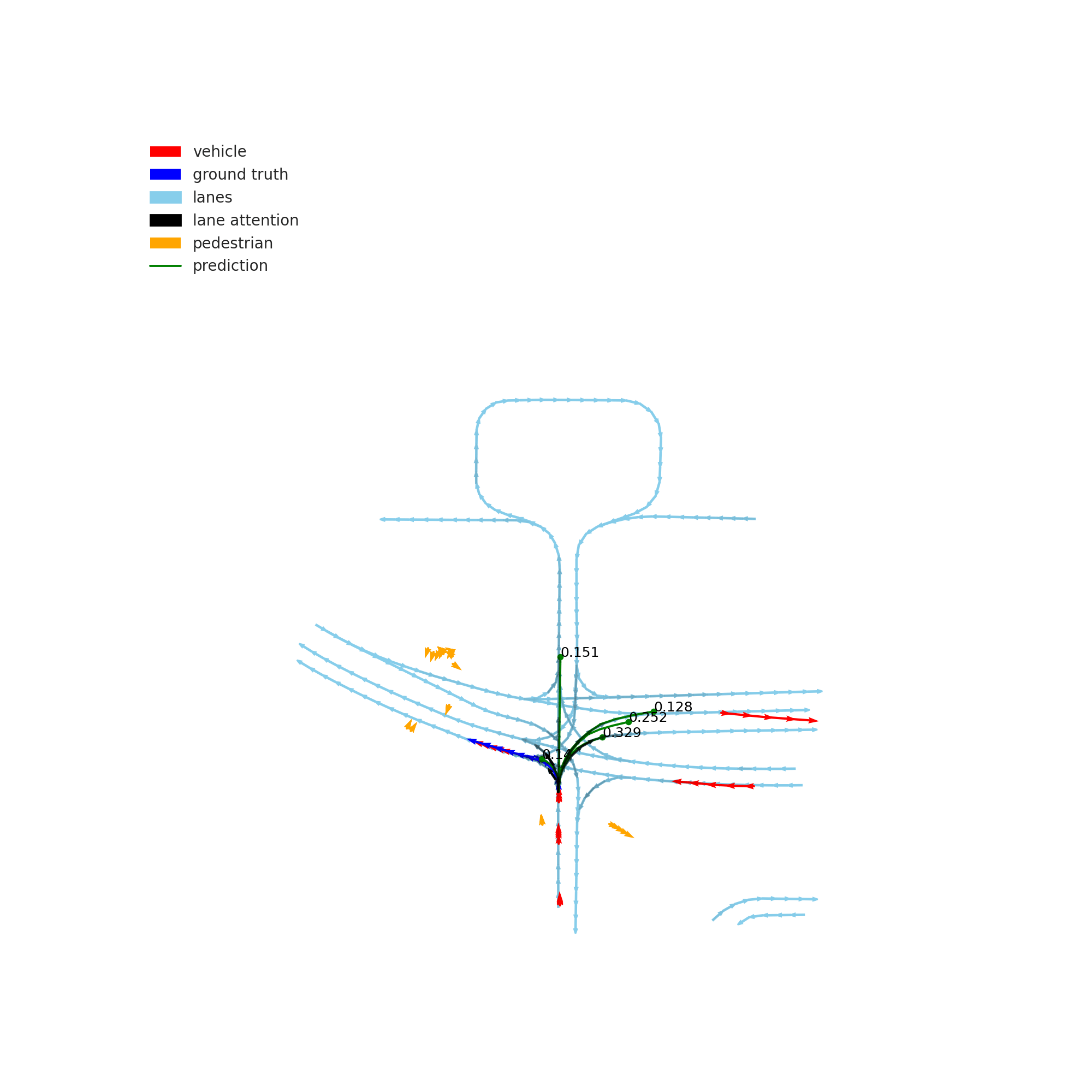}
        % \caption{Final Trajectories}
    \end{subfigure}
    % \begin{subfigure}[t]{0.4\linewidth}
    %     \includegraphics[height=4cm,trim={320 170 225 315},clip]{images_results/6f902cccc24d49a2b91d1d7f00047f83_6891260ad7c84844b5e74faf17303fff_minADE5_5.44_2-min.png}
    %     % \caption{Final Trajectories}
    % \end{subfigure}
    \begin{subfigure}[t]{0.4\linewidth}
        \includegraphics[height=4cm,trim={320 160 260 330},clip]{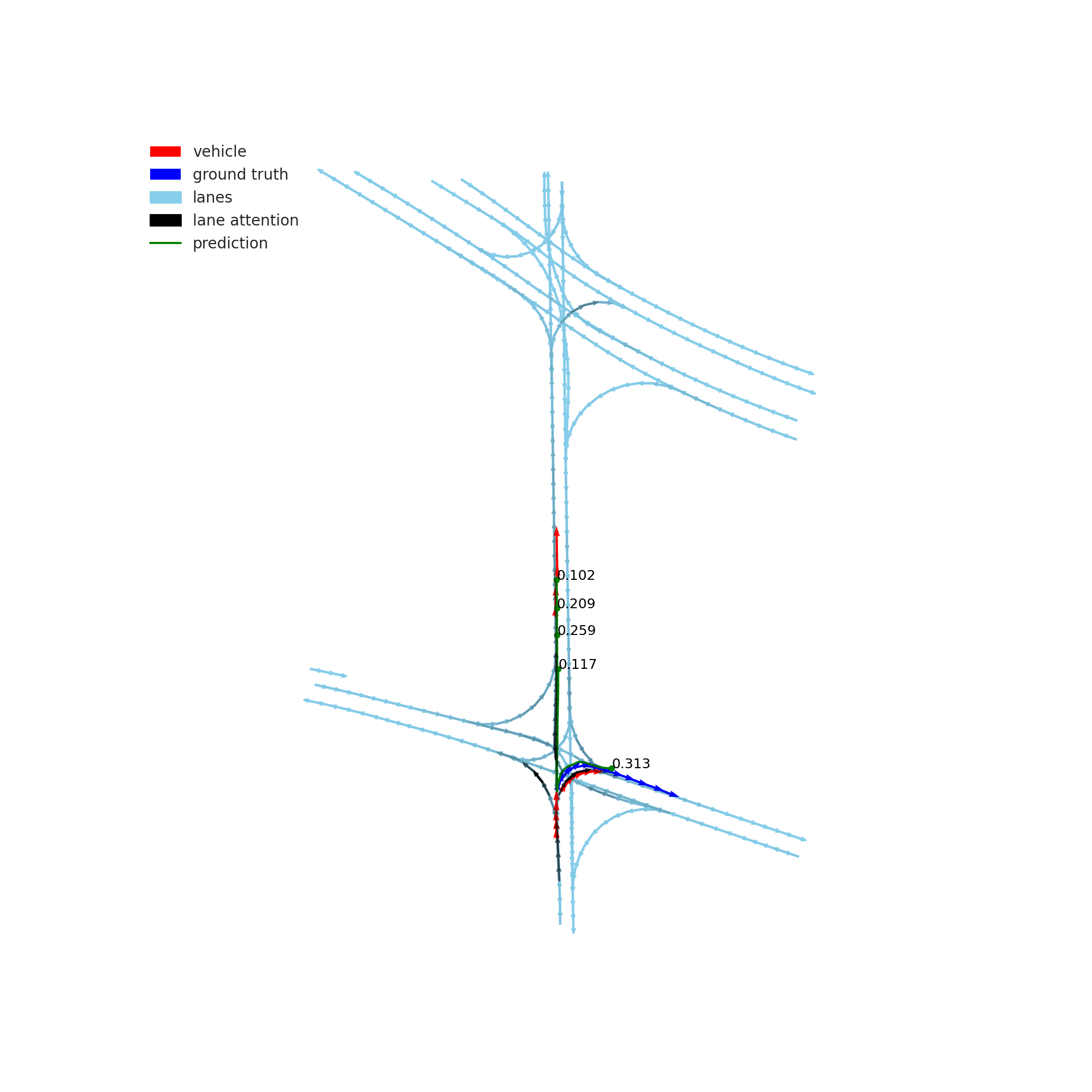}
        % \caption{Final Trajectories}
    \end{subfigure}
    \begin{tikzpicture}[remember picture,overlay]
      \node at (0, 3.3)      {\includegraphics[width=0.23\linewidth,trim={110 500 450 85},clip]{images_results/2706cc4eb61844a1a5c7a5cb766ffc2e_37ab2d88d07846ef96d5227e9c5a15d7_minADE5_5.01_0-min.png}};
    \end{tikzpicture}
    \caption{An illustration of predicted trajectories (green) with mode probabilities for the target agent at the final output layer. These samples are selected out of the 100 worst predictions based on minADE\textsubscript{5}.}
    \label{fig:nuScenes_worst}    
\end{figure}

We additionally investigate the worst reported prediction of LMFormer based on the minADE\textsubscript{5} metric (see Figure \ref{fig:nuScenes_worst}). Interestingly, even though the network generates diverse predictions, the predicted velocity for the mode following the ground truth path can be suboptimal, resulting in a bad minADE\textsubscript{k}. Thus, we hypothesize that the performance of the motion prediction modules can be improved by separately predicting velocity profiles and future paths. The work by Sun et al. \cite{sun2024semanticformer} also reports a similar observation, i.e., if the prediction modules have access to the ground truth velocity, the performance of the models improves significantly. Given this evidence, the direction of separation between velocity and path prediction should be explored in future work. 

\begin{figure}
    \centering
    \begin{subfigure}[t]{0.3\linewidth}
        \includegraphics[height=3cm,trim={280 160 300 370},clip]{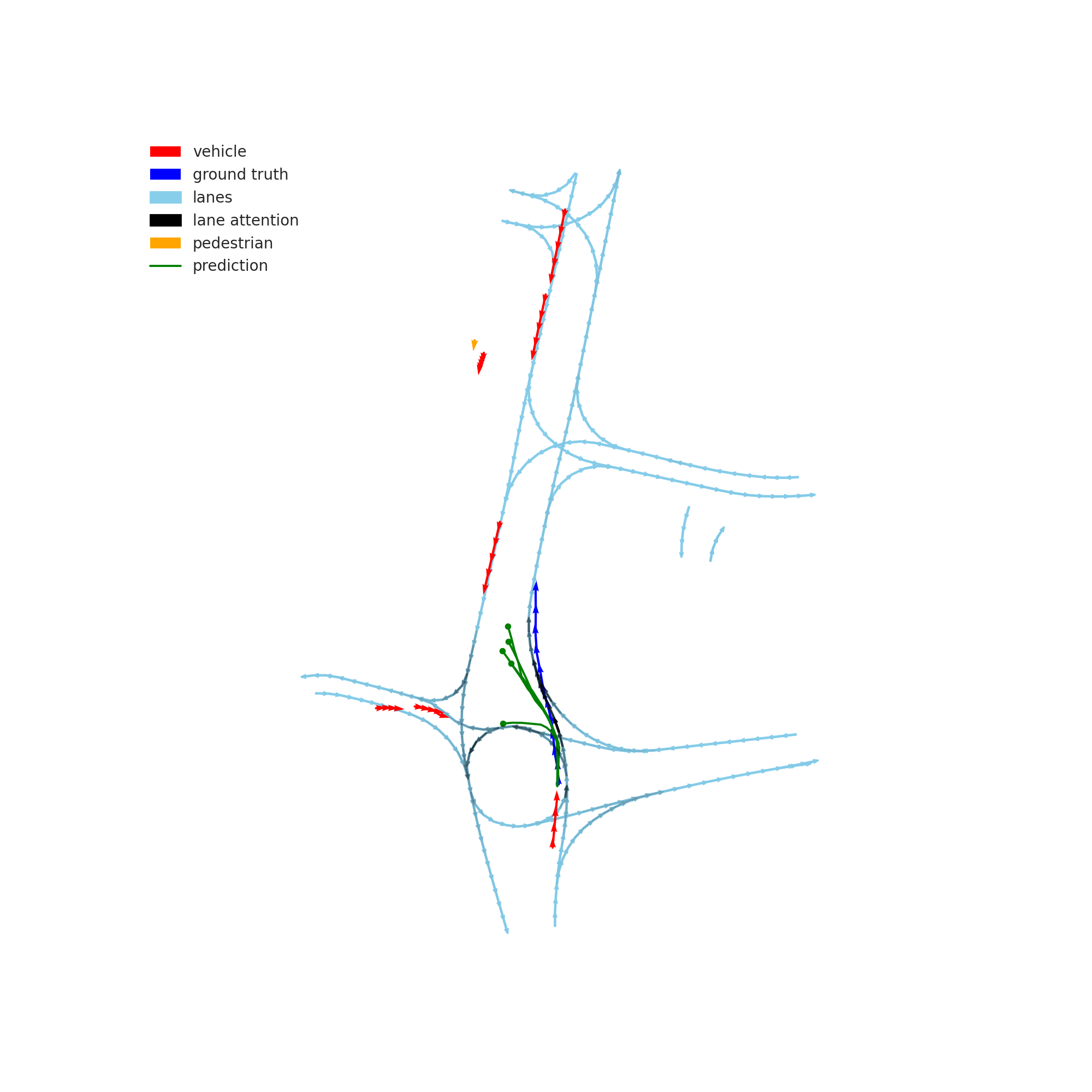}
    \end{subfigure}
    \begin{subfigure}[t]{0.3\linewidth}
        \includegraphics[height=3cm,trim={280 160 300 370},clip]{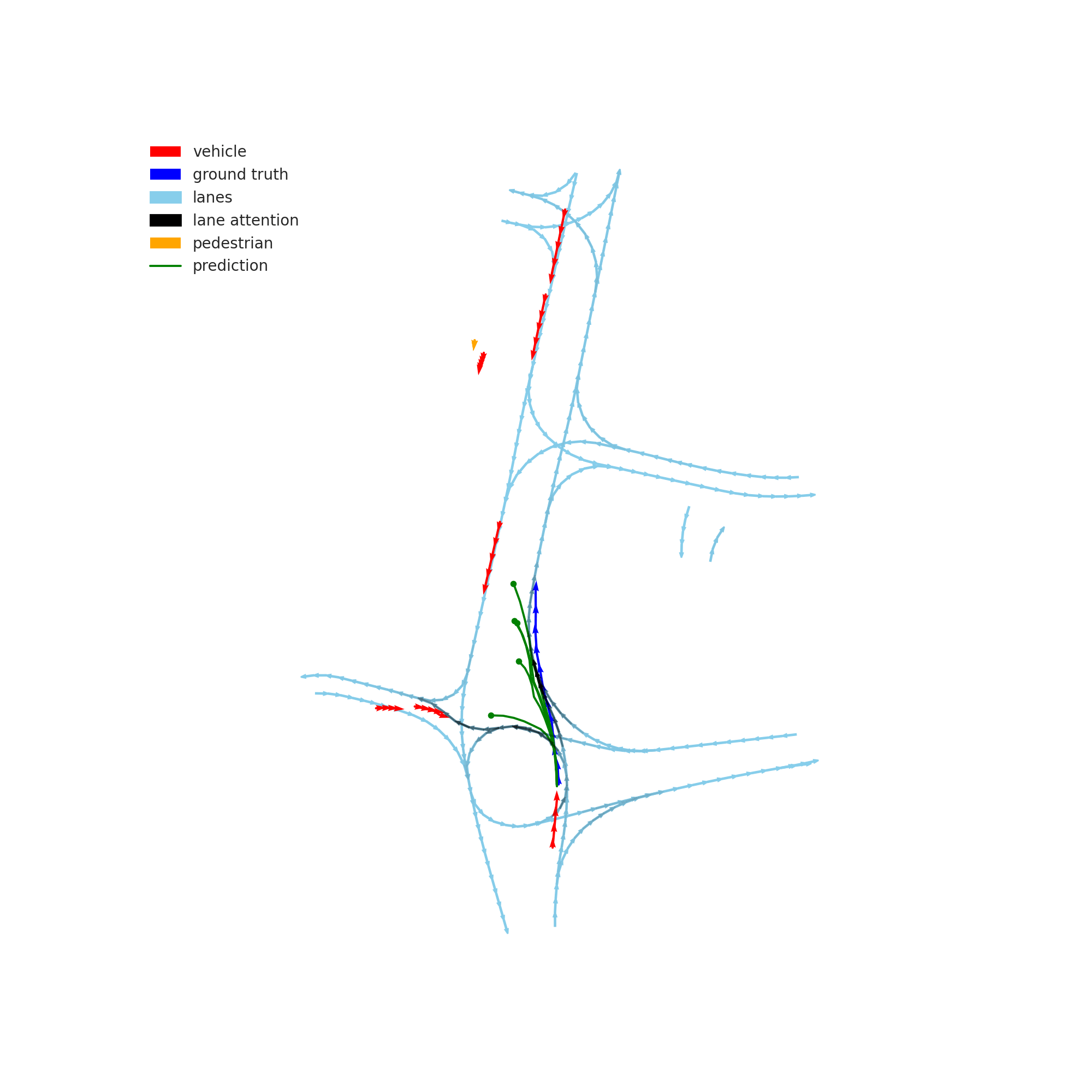}
    \end{subfigure}
    \begin{subfigure}[t]{0.3\linewidth}
        \includegraphics[height=3cm,trim={280 160 300 370},clip]{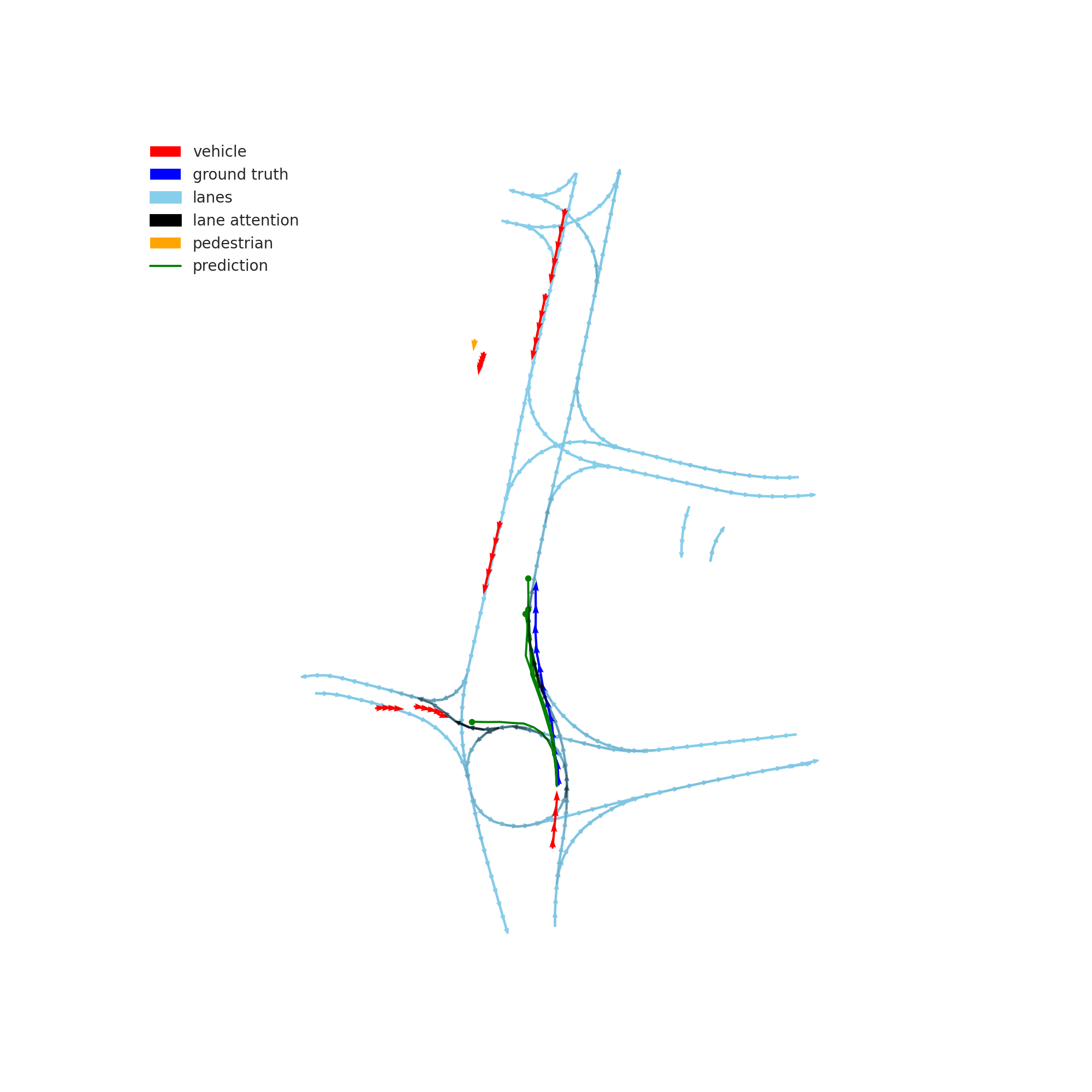}
    \end{subfigure}
    \begin{subfigure}[t]{0.3\linewidth}
        \includegraphics[height=3cm,trim={280 170 300 380},clip]{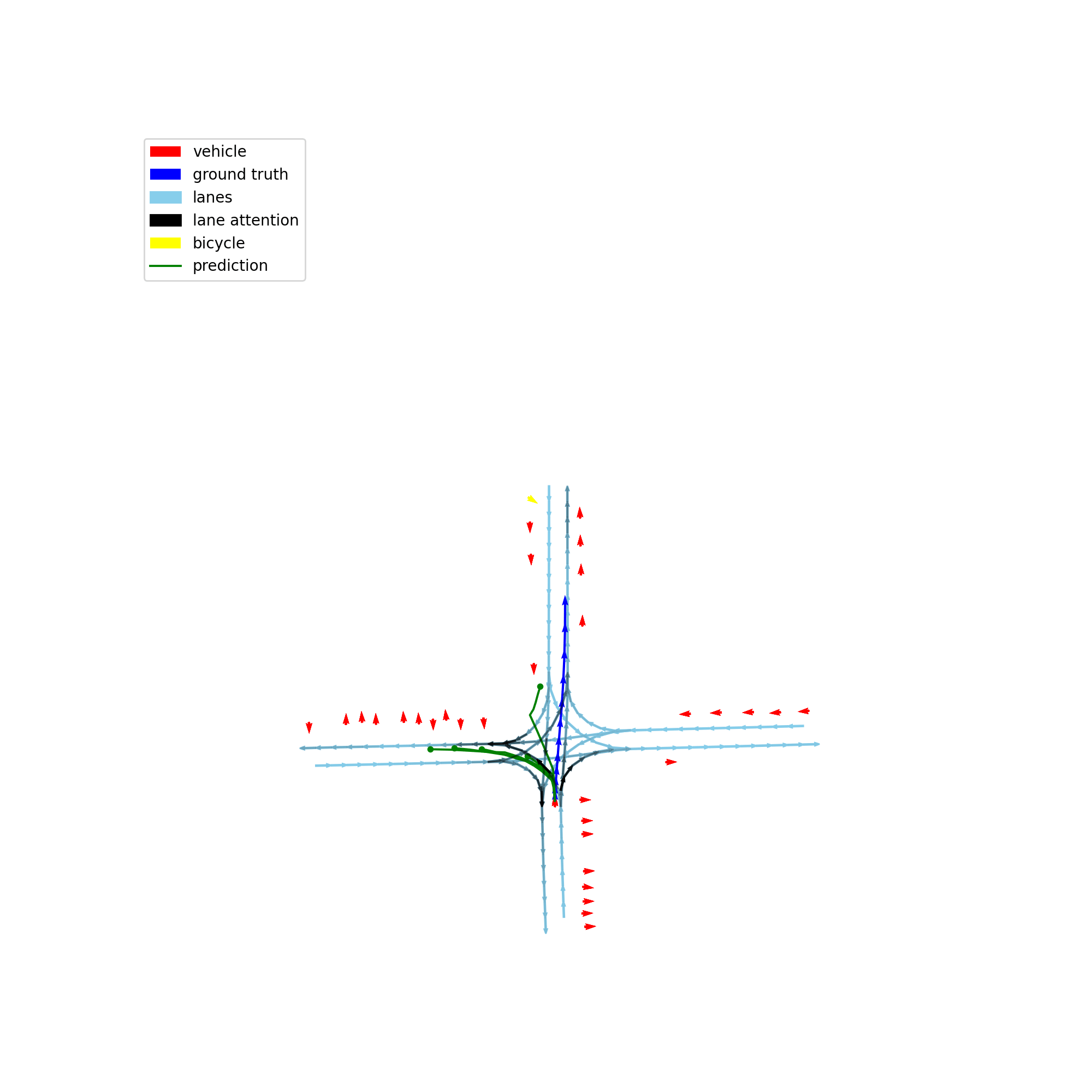}
        \caption{Initial}
    \end{subfigure}
    \begin{subfigure}[t]{0.3\linewidth}
        \includegraphics[height=3cm,trim={280 170 300 380},clip]{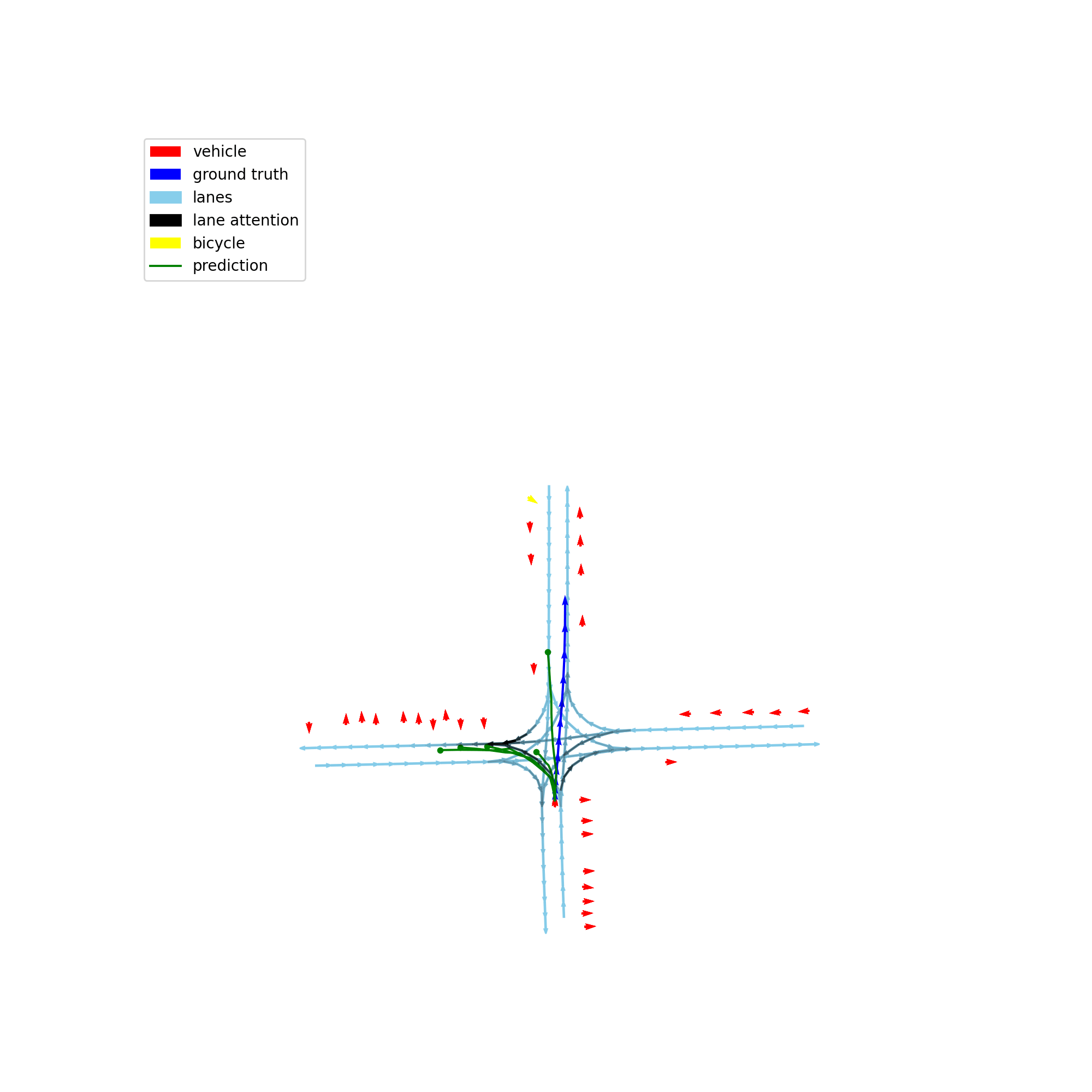}
        \caption{Refinement}
    \end{subfigure}
    \begin{subfigure}[t]{0.3\linewidth}
        \includegraphics[height=3cm,trim={280 170 300 380},clip]{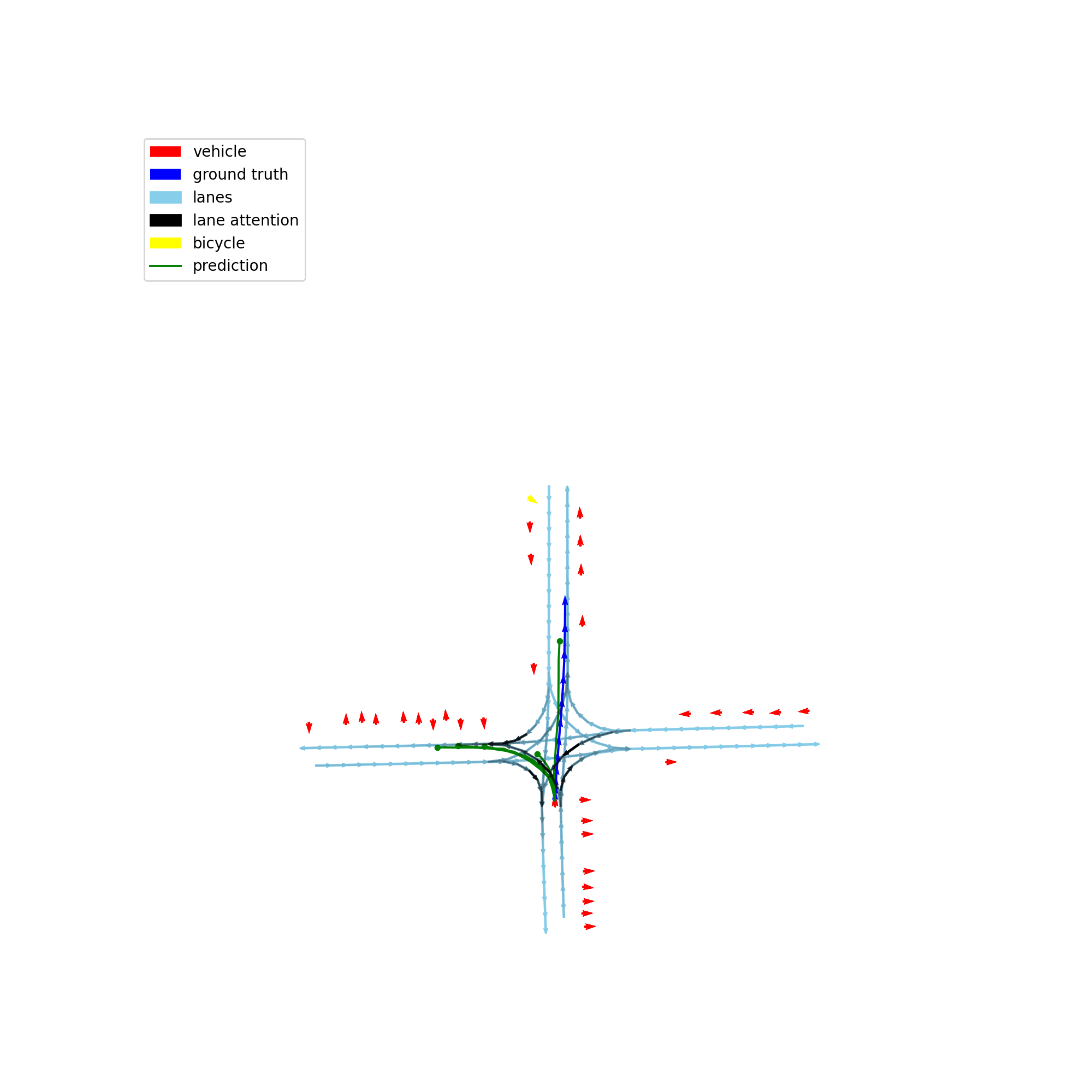}
        \caption{Final}
    \end{subfigure}
    % \begin{tikzpicture}[remember picture,overlay]
    %  \node at (0,3)      {\includegraphics[width=0.2\linewidth,trim={110 500 450 85},clip]{images_results/2706cc4eb61844a1a5c7a5cb766ffc2e_37ab2d88d07846ef96d5227e9c5a15d7_minADE5_5.01_0-min.png}};
    % \end{tikzpicture}
    \caption{Visualization of predicted trajectories (green) for a target agent at an unseen intersection during training. The model, trained on the Deep Scenario dataset, refines its predictions progressively from (a) to (c), achieving better alignment with the lane structure. However, as shown in the second row, a limitation of our approach becomes evident: despite the attention maps indicating that the network should predict right-turn trajectories, it fails to generate them.}
    \label{fig:cross_data_qual}    
\end{figure}

We also test the generalization capability of LMFormer on the Deep Scenario dataset \cite{lu2023deepscenario}. To this end, we train the network on 55,000 training samples, generated from four different intersections' recordings ('Unparalleled Frankfurt', 'Trendy Renningen', Fortunate Karlsruhe', 'Stunning Stuttgart') in the dataset. We conducted the qualitative testing for this model checkpoint on the samples of unobserved intersections during training (refer Figure \ref{fig:cross_data_qual}). We observe that in certain cases, LMFormer fails to predict all plausible trajectories. We hypothesize that this limitation is due to an unbalanced data distribution, where some maneuvers occur less frequently in the training data. This suggests that additional mechanisms, such as explicit trajectory diversity constraints or targeted data augmentation, may be required to improve the robustness of multimodal prediction, but we leave these improvements for future work.

\begin{table}[b]
    \centering
    \caption{LMFormer Cross Datataset performance validated on nuScenes validation split}
    \label{tab:train_ds_eval_nuScenes}
    \resizebox{\linewidth}{!}{
    \begin{tabular}{c c|c c c|c}
    \hline
    \multicolumn{2}{c|}{Training Dataset} & & & & \\
    Deep Scenario & nuScenes & minADE\textsubscript{5}$\downarrow$ & MR\textsubscript{5}$\downarrow$ & minFDE\textsubscript{5}$\downarrow$ & OffRoad$\downarrow$ \\
    \hline
    \checkmark & - & 1.54 & 0.57 & 2.86 & 0.02\\
    - & \checkmark & 1.13 & 0.48 & 2.13 & 0.01 \\
    \checkmark &\checkmark & 1.10 & 0.46 & 2.07 & 0.01\\
    \hline
    % \checkmark & - & 1.56 & 0.57 & 3.02 & 0.02 \\
    % - & \checkmark & 1.13 & 0.47 & 2.14 & 0.01 \\
    % \checkmark &\checkmark & 1.10 & 0.44 & 2.09 & 0.01\\
    \end{tabular}
    }
\end{table}

Furthermore, a quantitative evaluation of this model checkpoint, which is trained on Deep Scenario, is done on the nuScenes validation split to assess its cross-dataset generalization capabilities. The corresponding results are presented in Table \ref{tab:train_ds_eval_nuScenes}. It can be observed that the performance of this model checkpoint is significantly worse in comparison to the one trained on the nuScenes training set. In our analysis of this cross-dataset performance degradation, we identified a systematic underestimation of the predicted velocity in the case of Deep Scenario as compared to nuScenes. We attribute this discrepancy to the distributional differences between the datasets: Deep Scenario captures mainly slow-moving vehicles in intersection-heavy environments, while nuScenes provides a more diverse range of driving dynamics. To test this hypothesis, we train LMFormer on a combined dataset, containing training samples from both the nuScenes and Deep Scenario training sets. The corresponding qualitative results are shown in Table \ref{tab:train_ds_eval_nuScenes}. We observe that this model checkpoint outperforms the others trained individually on Deep Scenario and nuScenes training sets. This indicates that the network can learn to generalize better with more diverse datasets and can overcome limitations present in individual datasets, as previously reported in UniTraj \cite{feng2024unitraj}. It is important to note that the network architecture as well as the number of parameters were kept constant in the cross-dataset evaluations. 

\begin{figure}
    \centering
    \begin{subfigure}[t]{0.4\linewidth}
        \includegraphics[height=4cm,trim={270 100 250 350},clip]{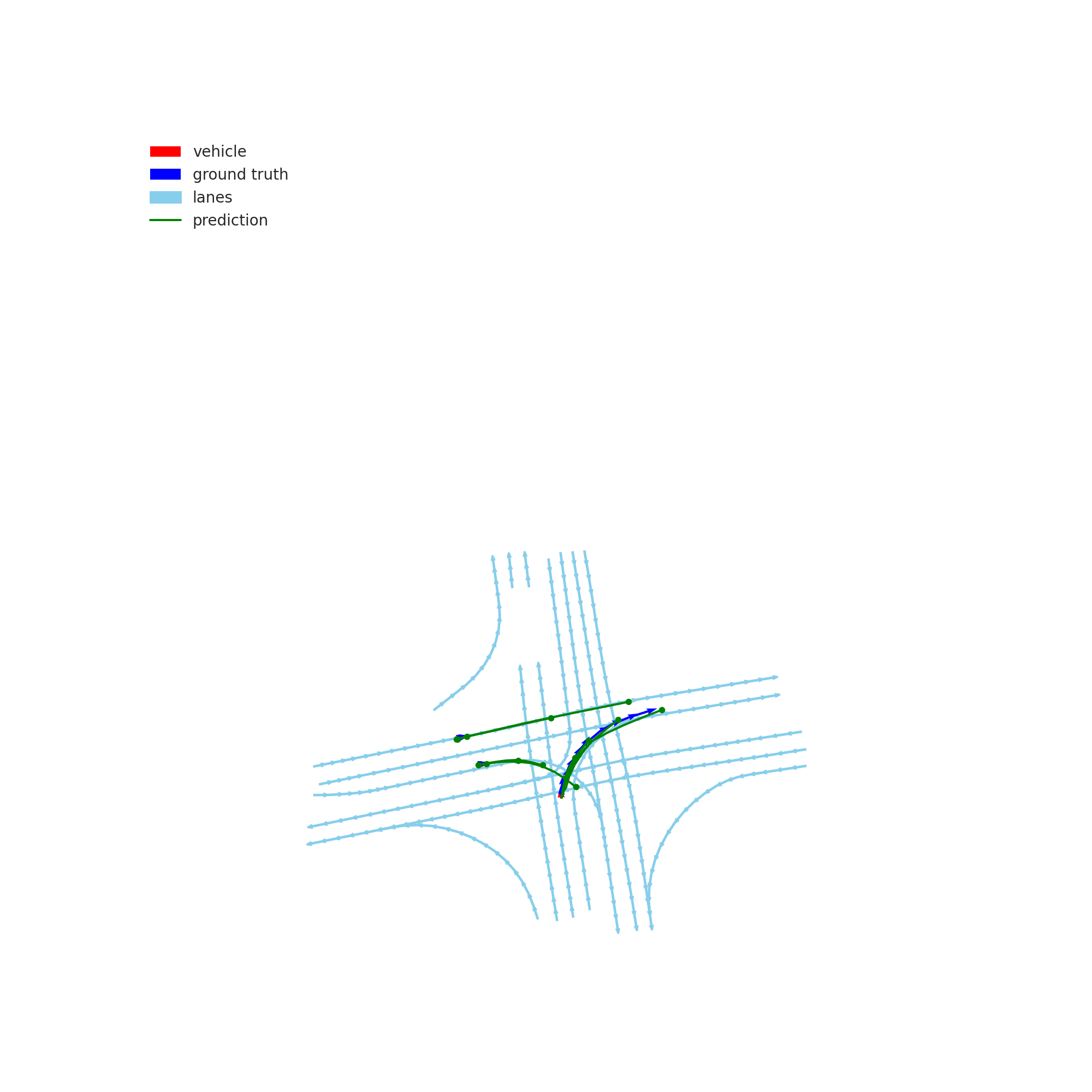}
        % \caption{Final Trajectories}
    \end{subfigure}
    \begin{subfigure}[t]{0.2\linewidth}
        \includegraphics[height=4cm,trim={280 130 240 300},clip]{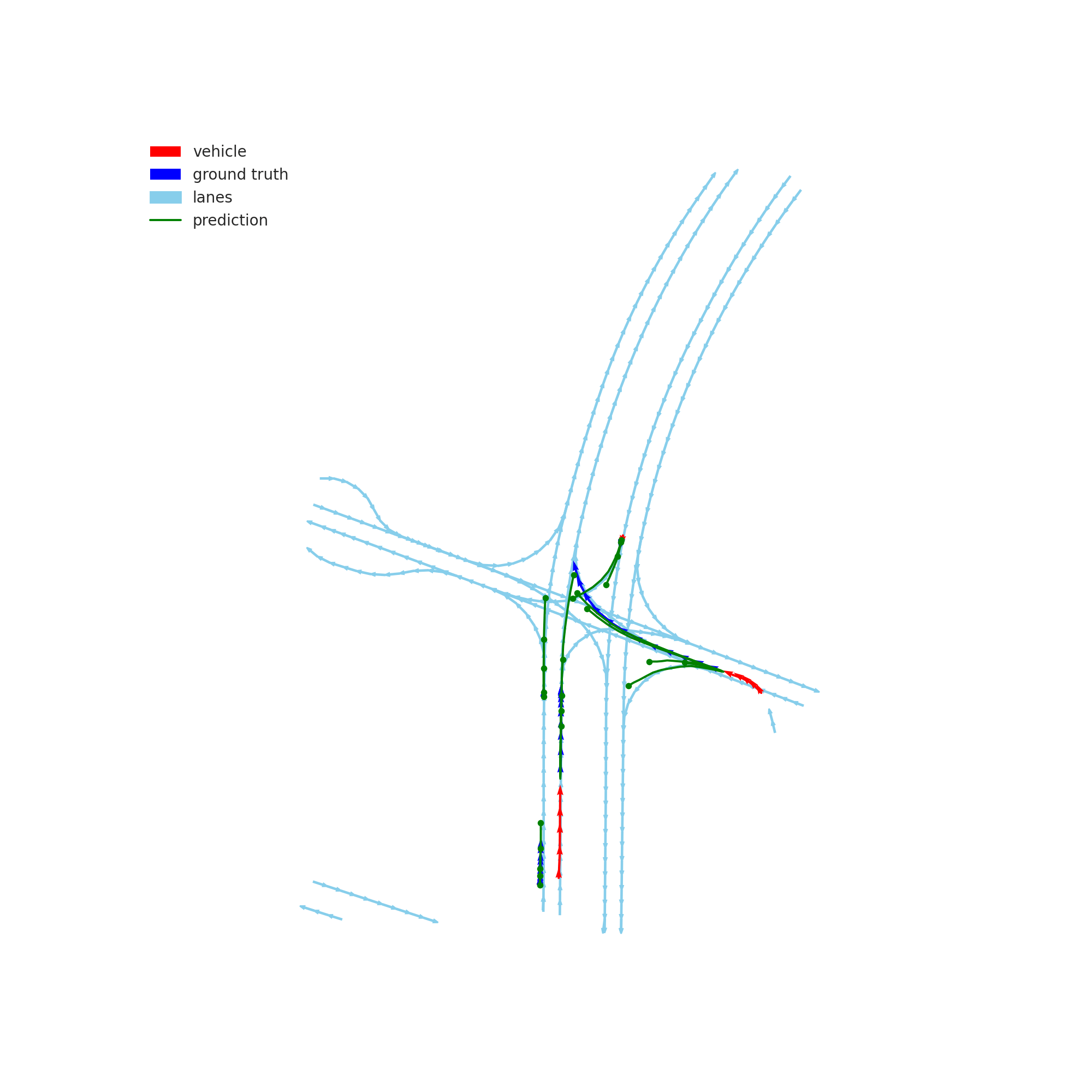}
        % \caption{Final Trajectories}
    \end{subfigure}
    % \begin{tikzpicture}[remember picture,overlay]
    %  \node at (0,7)      {\includegraphics[width=0.2\linewidth,trim={110 500 450 30},clip]{images_results/2706cc4eb61844a1a5c7a5cb766ffc2e_37ab2d88d07846ef96d5227e9c5a15d7_minADE5_5.01_0-min.png}};
    % \end{tikzpicture}
    \caption{A depiction of marginal scene prediction. It illustrates the capability that LMFormer can be extended to multi-agent prediction.}
    \label{fig:scene_prediction}    
\end{figure}

In addition, we also illustrate the capability of LMFormer to predict multiple agents in parallel. Figure \ref{fig:scene_prediction} shows a marginal scene prediction output, where the trajectories are predicted in a single forward pass for all vehicles in the scene. 

\subsection{Ablations}\label{subsection:ablations}
In our ablation study, we validate the two key components of our approach: (1) the refinement strategy and (2) the long-range lane encoder. To evaluate the impact of refinement, we remove the additional regression loss introduced by the refinement layers. Similarly, we remove the lane self-attention module to evaluate the role of long-range lane interactions within the Map Encoder. The results of the ablation are presented in Table \ref{tab:ablation}. These results show that removing either component leads to a degradation in performance, with the degradation being slightly more pronounced when refinement losses are omitted. Overall, the ablation study confirms that both proposed components contribute significantly to SOTA performance.

\begin{table}[h]
    \centering
    \caption{Ablation Study conducted on the nuScenes val split}
    \label{tab:ablation}
    \resizebox{\linewidth}{!}{
    \begin{tabular}{c c|c c c|c}
    \hline
    \makecell{Lane \\ Self Attention} & Refinement & minADE\textsubscript{5}$\downarrow$ & MR\textsubscript{5}$\downarrow$ & minFDE\textsubscript{5}$\downarrow$ & OffRoad$\downarrow$ \\
    \hline
    \checkmark & \checkmark & 1.13 & 0.48 & 2.13 & 0.01 \\
     \checkmark & - & 1.16 & 0.51 & 2.20 & 0.01 \\
    - & \checkmark & 1.15 & 0.49 & 2.19 & 0.01 \\
    \hline
    %  \checkmark & \checkmark & 1.13 & 0.47 & 2.14 & 0.01 \\
    %  \checkmark & - & 1.14 & 0.48 & 2.16 & 0.01 \\
    %  \checkmark & - & 1.15 & 0.48 & 2.16 & 0.01 \\
    % - & \checkmark & 1.15 & 0.47 & 2.19 & 0.01 \\
    % - & \checkmark & 1.15 & 0.47 & 2.18 & 0.01 \\
    \end{tabular}
    }
\end{table}
\section{Conclusion}
The paper introduces a transformer-based motion prediction module and tackles several important challenges such as trajectory refinement and lane attention. Our work establishes that lanes are the most important components in the static context for prediction tasks. Although LMFormer quantitatively achieves SOTA performance, the qualitative results show that further improvements are needed to address robustness in prediction diversity. We also note that the current evaluation benchmarks are limited in scope and could benefit from additional metrics for velocity profiles and predicted paths. Our cross-dataset evaluation highlights a potential limitation of training on intersection-centric data when generalizing to broader urban driving scenarios, but this issue could be overcome by combining samples from individual datasets.
{
    \small
    \bibliographystyle{ieeenat_fullname}
    \bibliography{main}
}


\end{document}